\documentclass[runningheads]{llncs}
\usepackage{xcolor}
\usepackage{bbding}
\usepackage{amssymb}
\usepackage{dsfont}

\usepackage{tikz}
\usepackage{comment}
\usepackage{amsmath} 
\usepackage{color}

\usepackage{booktabs}
\usepackage{multirow}
\usepackage{makecell}
\usepackage[perpage]{footmisc}

\usepackage{float}
\usepackage{soul}
\usepackage{caption}
\usepackage{subcaption}
\usepackage{graphicx}
\usepackage[pagebackref,breaklinks,colorlinks]{hyperref}
\usepackage[width=122mm,left=12mm,paperwidth=146mm,height=193mm,top=12mm,paperheight=217mm]{geometry}

\usepackage[capitalize]{cleveref}
\usepackage{cite}
\usepackage{xcolor, colortbl}
\usepackage[normalem]{ulem}
\usepackage[accsupp]{axessibility}  
\newcommand{\beginsupplement}{%
        \setcounter{table}{0}
        \renewcommand{\thetable}{S\arabic{table}}%
        \setcounter{figure}{0}
        \renewcommand{\thesection}{\Alph{section}}
     }

\newcommand\samethanks[1][\value{footnote}]{\footnotemark[#1]}

\begin{document}
\pagestyle{headings}
\mainmatter
\def\ECCVSubNumber{6295}  

\title{Pixel-wise Energy-biased Abstention Learning for Anomaly Segmentation on Complex Urban Driving Scenes } 


\titlerunning{PEBAL}
%
\author{Yu Tian \inst{1}$\thanks{First two authors contributed equally to this work. GP is the corresponding author.}$,
Yuyuan Liu \inst{1}${\samethanks}$, 
Guansong Pang \inst{2}, 
Fengbei Liu \inst{1}, 
Yuanhong Chen \inst{1}, and
Gustavo Carneiro \inst{1}}

\authorrunning{Y. Tian et al.}
%
\institute{Australian Institute for Machine Learning, University of Adelaide \and 
 Singapore Management University 
}
\maketitle

\vspace{-20pt}

\begin{abstract}

State-of-the-art (SOTA) anomaly segmentation approaches on complex urban driving scenes explore pixel-wise classification uncertainty learned from outlier exposure, or external reconstruction models.
However, 
previous uncertainty approaches that directly associate high uncertainty to anomaly may sometimes lead to incorrect anomaly predictions, and 
external reconstruction models tend to be too inefficient for real-time self-driving embedded systems. 
In this paper, we propose a new anomaly segmentation method, named pixel-wise energy-biased abstention learning (PEBAL), that explores pixel-wise abstention learning (AL) with a model that 
learns 
an adaptive pixel-level anomaly class, and an energy-based model (EBM) that learns inlier pixel distribution.
More specifically, PEBAL is based on a non-trivial joint training of EBM and AL, where EBM is trained to output high-energy for anomaly pixels (from outlier exposure) and AL is trained such that these high-energy pixels receive adaptive low penalty for being included to
the anomaly class.
We extensively evaluate PEBAL against the SOTA and show that it achieves the best performance across four benchmarks. 
Code is available at \url{https://github.com/tianyu0207/PEBAL}.

\end{abstract}

\vspace{-30pt}
\section{Introduction}
\label{sec:intro}

\begin{figure*}[t!]
    \centering
    \includegraphics[width=1.\textwidth]{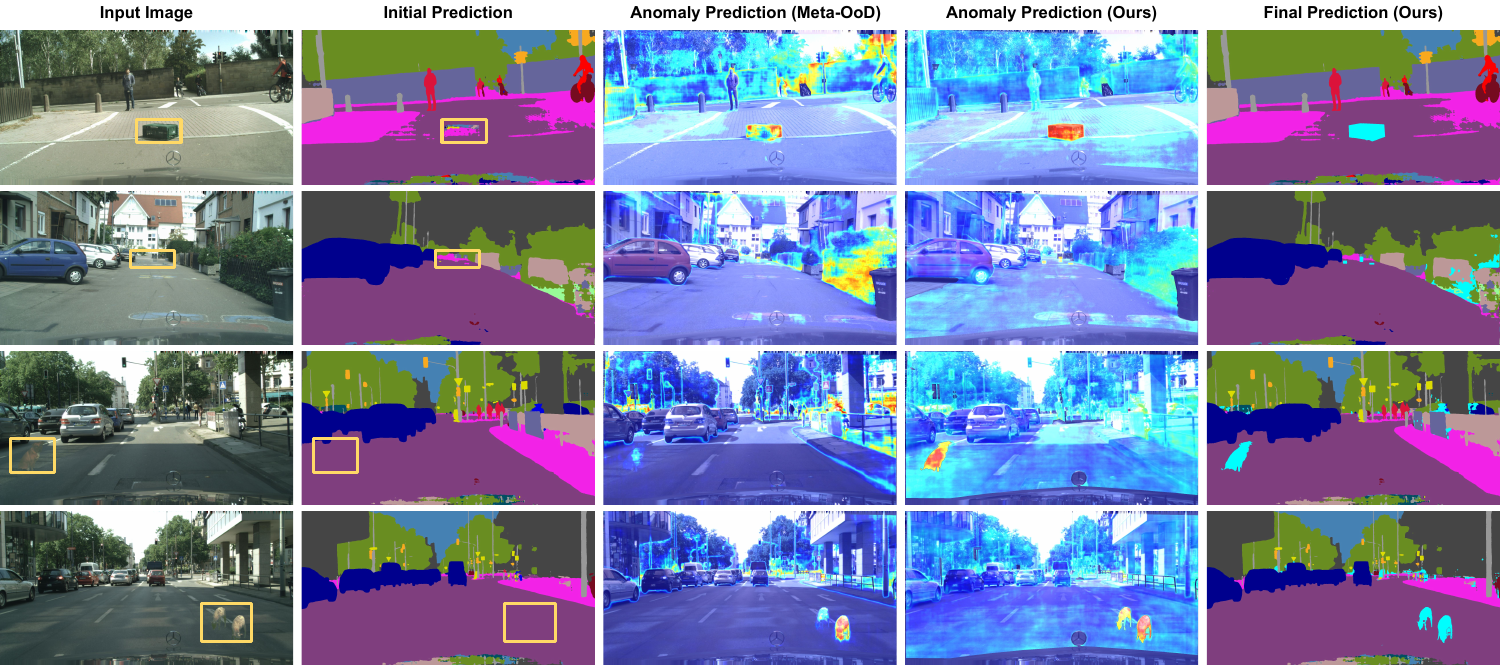}
    \vspace{-12pt}
    \captionof{figure}{\textbf{Anomaly segmentation overview}. 
    From the \textbf{input image} (anomaly highlighted with a yellow box), the \textbf{initial prediction} shows the original segmentation results with anomalies classified as a one of the pre-defined inlier classes.
    \textbf{Anomaly predictions} by the previous SOTA \textbf{Meta-OoD~\cite{chan2021entropy}} and \textbf{our} method
    show an anomaly map with high scores (in yellow and red) for anomalous pixels, where our approach shows less false positive and false negative detections.
    Consequently, our method can detect 
    small and distant anomalies (row 2)  and blurry/unclear anomalies (rows 1, 3, 4) more accurately than Meta-OoD~\cite{chan2021entropy}. 
    In our \textbf{final prediction}, anomalous pixels are coloured as cyan.  \textbf{Some anomalies are small and blurred (e.g.,  row 2), so please zoom in the PDF for better visualisation.
    }
    }
    \vspace{-12pt}
    \label{fig:intro}
\end{figure*}

Recent advances in semantic segmentation have shown tremendous improvements on complex urban driving scenes~\cite{lateef2019survey,liu2022perturbed,ouali2020semi, zou2020pseudoseg, french2019semi,chen2021semi}. Despite the accurate predictions on the inlier classes, 
the model fails to properly recognise  anomalous objects that deviate from the training inlier distribution (col. 2 of Fig.~\ref{fig:intro}).
Addressing such failure cases is crucial to road safety for autonomous driving vehicles. 
For example, anomalies can be represented by unexpected objects in the middle of the road, such as
a large rock or an unexpected animal that can be incorrectly predicted as a part of the road class, leading to potentially fatal traffic collisions.

Current methods~\cite{mukhoti2018evaluating,bevandic2019simultaneous,jung2021standardized,di2021pixel,chan2021entropy,xia2020synthesize,lis2019detecting,blum2019fishyscapes} to detect and segment anomalous objects in complex urban driving scenes tend to depend on classification uncertainty or image reconstruction. 
The association of high classification uncertainty with anomaly is intuitive, but it has a few caveats.  
For instance, classification uncertainty happens when samples are close to classification decision boundaries, but there is no guarantee that all anomalies will be close to classification boundaries.
Furthermore, samples close to classification boundaries may not be anomalies at all, but just hard inlier samples. 
Hence, these uncertainty based methods may detect a large number of false positive and false negative anomalies.
For example, Fig.~\ref{fig:intro} shows that the previous SOTA Meta-OoD~\cite{chan2021entropy} misses important anomalous pixels (all rows), while misclassifying anomalies (e.g., vegetation in rows 1, 2, 3), even with the use of the outlier exposure (OE) strategy~\cite{hendrycks2018deep}. In fact, the OE strategy maximises the uncertainty for proxy anomalies, which can cause the model to be more uncertain for all inlier classes and detect false positive anomalies (e.g., Meta-OoD mis-classifies trees or bush with high anomaly scores -- Fig.~\ref{fig:intro} col 4). 
Reconstruction methods~\cite{di2021pixel,xia2020synthesize} add an extra network to reconstruct the input images from the estimated segmentation, where differences are assumed to be anomalous. Not only does this approach depend on accurate segmentation results for precise reconstruction, but they also require an extra reconstruction network that is hard to train and inefficient to run in real-time self-driving embedded systems. 
Moreover, reconstruction methods that rely on a discrepancy 
module require re-training whenever the inlier segmentation model changes due to input distribution shift~\cite{di2021pixel}, limiting their applicability in real-world systems.
 Furthermore, previous approaches~\cite{chan2021entropy,di2021pixel,bevandic2019simultaneous,lis2019detecting,jung2021standardized,grcic2021dense} ignore a couple of important constraints for anomaly segmentation, namely smoothness (e.g., Meta-OoD fails to classify neighbouring anomaly pixels in Fig.~\ref{fig:intro}, rows 1, 4) and sparsity (e.g., Meta-OoD incorrectly detects a large number of anomalous pixels--see yellow and red regions in Fig.~\ref{fig:intro}, rows 1, 2, 3).
Another common issue shared by previous methods~\cite{chan2021entropy,bevandic2019simultaneous,lis2019detecting} is that they usually rely on the re-training of the entire network for OE, which is inefficient and can also bias the classification towards outliers.

In this paper, we propose a new anomaly segmentation method, the pixel-wise energy-biased abstention learning (PEBAL), that directly learns a pixel-level anomaly class, in addition to the pre-defined inlier classes, to reject/abstain anomalous pixels that are dissimilar to any of the inlier classes. It is achieved by a joint optimisation of a novel pixel-wise anomaly abstention learning (PAL) and an energy based model (EBM)~\cite{grathwohl2019your,liu2020energy,lecun2006tutorial}. 
Particularly, abstention learning (AL)~\cite{liu2019deep} was originally developed to learn an image-level anomaly class, which is significantly challenged by the pixel-wise anomaly segmentation task that requires pixel-level anomaly class learning. This is because the original AL model treats all pixel inputs equally with a single pre-defined fixed penalty factor to regularise the classification of anomalous pixels, while adaptive penalties are typically required for different pixels in a complex driving scene, e.g., pixels in small (distant) objects vs. large (near) objects, or centred pixels vs fringe pixels of objects. PEBAL is designed to address this issue by learning adaptive pixel-wise energy-based penalties, which automatically decreases the penalty for pixels that are likely to be anomalies.
Hence, our model does not explore previously proposed uncertainty measures (e.g., entropy or softmax criteria) or image reconstruction, and instead, for the first time, explicitly learns a new pixel-wise anomaly class. The learned penalty factors are jointly optimised with EBM, resulting in a mutually beneficial optimisation of anomaly and inlier segmentation. 
Additionally, we impose smoothness and sparsity constraints to the learning of the anomaly segmentation by PEBAL, incorporating local and global dependencies into the pixel-wise penalty estimation and anomaly score learning.
Finally, the training of PEBAL is  efficient given that we only need to fine-tune the last block of the segmentation model to achieve accurate inference.
To summarise, our contributions are the following: 
 \begin{itemize}
 \item We propose the pixel-wise energy-biased abstention learning (PEBAL) that jointly optimises a novel pixel-wise anomaly abstention learning (PAL) and energy-based models (EBM) to learn adaptive pixel-level anomalies. PEBAL mutually reinforces PAL and EBM in detecting anomalies, enabling accurate segmentation of anomalous pixels without compromising the segmentation of inlier pixels (cols. 4,5 of Fig.~\ref{fig:intro}). 
 \item We introduce a new pixel-wise energy-biased penalty estimation, which can learn adaptive energy-based penalties to highly varying pixels in a complex driving scene, allowing a robust detection of small/distant and blurry anomalous objects (Fig.~\ref{fig:intro} row 2).  
 \item We further refine our PEBAL training, using a novel smoothness and sparsity regularisation on anomaly scores to consider the local and global dependencies of the pixels, enabling the reduction of false positive/negative anomaly predictions.
 \end{itemize}

We validate our approach on Fishyscapes leaderboard~\cite{blum2019fishyscapes}, and achieve SOTA classification accuracy on all relevant benchmarks. We also achieve the best classification results on LostandFound~\cite{pinggera2016lost} and Road Anomaly~\cite{lis2019detecting} test sets, significantly surpassing other competing methods. 

\section{Related work}
\label{sec:related}
\noindent\textbf{Uncertainty-based Anomaly Segmentation.}
Early uncertainty-based methods~\cite{lee2017training,liang2017enhancing,hendrycks2016baseline,tian2021weakly} focused on the estimation of image-level anomalies, but 
they tended to misclassify object boundaries as anomalies~\cite{jung2021standardized}. 
Jung et al.~\cite{jung2021standardized} mitigate this issue by iteratively replacing false anomalous boundary pixels with neighbouring non-boundary pixels that have low anomaly score. 
In~\cite{kendall2017uncertainties,lakshminarayanan2016simple,mukhoti2018evaluating}, the boundary issue was tackled with a pixel-wise uncertainty estimated with MC dropout, but they showed a low pixel-wise anomaly detection accuracy~\cite{lis2019detecting}. Without fine-tuning using a proxy outlier dataset, uncertainty estimation may not be accurate enough to detect anomalies and can predict high uncertainty for challenging inliers or low uncertainty for outliers due to overconfident misclassification. 

\noindent\textbf{Reconstruction-based Anomaly Segmentation.}
Anomalies can also be segmented from the errors between the input image and its reconstruction obtained from its predicted segmentation map~\cite{baur2018deep,creusot2015real,chen2021deep,haldimann2019not,lis2019detecting,xia2020synthesize,di2021pixel,vojir2021road,liu2020photoshopping,tian2021constrained}. 
Those approaches are challenged by the dependence on an accurate segmentation prediction, by the complexity of reconstruction models 
that usually require long training and inference processes, and also by the low quality of the reconstructed images.

\noindent\textbf{Anomaly Segmentation via Outlier Exposure.}
Hendrycks et al.~\cite{hendrycks2018deep}  propose the outlier exposure (OE) strategy that uses an auxiliary dataset of outliers that do not overlap with the real outliers/anomalies to improve the anomaly detection performance. 
This OE strategy uses outliers from ImageNet~\cite{bevandic2018discriminative,bevandic2019simultaneous,vandenhende2020revisiting,tian2020few}, void class of Cityscape~\cite{di2021pixel} or COCO~\cite{chan2021entropy}, where the expectation is that the model can generalise to unseen outliers.
Maximising uncertainty for outliers using the OE strategy can lead to a deterioration of the segmentation of inliers~\cite{bevandic2018discriminative,vandenhende2020revisiting}.
Another major drawback of OE methods is that they are trained using outlier images or objects without considering the fact that outliers are rare events that appear around inliers. Hence, the training contains a disproportionately high amount of outliers~\cite{chan2021entropy} that can bias the segmentation toward the anomaly class. We address this issue by respecting the anomaly detection assumption, where anomalous objects are rare, contribute to a small proportion of the training set, and appear around inliers.

\noindent\textbf{Abstention Learning.}
The abstention learning  mechanism~\cite{el2010foundations} adds a ``reserve" (i.e., anomaly) class that is predicted when the classification predictions for all inlier classes are not high enough. This method shows good performance in learning holistic image-level anomaly class with a single pre-defined penalty factor for the whole training set, but it fails to learn fine-grained pixel-level anomaly class as an adaptive pixel-wise penalty is required for highly varying pixel-level anomalies (see Table \ref{tab:ablation}). We address this issue by learning a novel pixel-wise energy-biased penalty estimator that is jointly trained with fine-grained abstention learning.
It is worth noting that differently from uncertainty-based methods~\cite{chan2021entropy,jung2021standardized,hendrycks2019scaling,blum2019fishyscapes} that assume anomaly even when the model is uncertain but confident, abstention learning requires all classes to have low confidence to predict the anomaly class.

\noindent\textbf{Energy-based Models.}
EBM is trained such that inlier training samples have low energy, whereas non-training outlier samples (i.e., anomalies) are expected to have high energy~\cite{lecun2006tutorial}.
This energy value can then be used to compute the probability of a sample to belong to the inlier distribution.
Recently, EBMs are being implemented with deep learning models~\cite{nijkamp2019learning,grathwohl2019your,liu2020energy}, and to learn them, it is necessary to compute the partition function, which is generally estimated with Markov Chain Monte Carlo (MCMC)~\cite{grathwohl2019your}, but this estimation cannot generate accurate high-resolution images.
Hence, we follow the simpler idea of estimating the energy score with the $logsumexp$ operator~\cite{grathwohl2019your,liu2020energy}, where we minimise the energy of inliers and use an OE  strategy~\cite{hendrycks2018deep} to maximise the energy of outliers. Hence, we do not need to compute the partition function.
\vspace{-10pt}

\section{Method}
\label{sec:method}

\begin{figure*}[ht!]
    \centering
    \includegraphics[width=.9\textwidth]{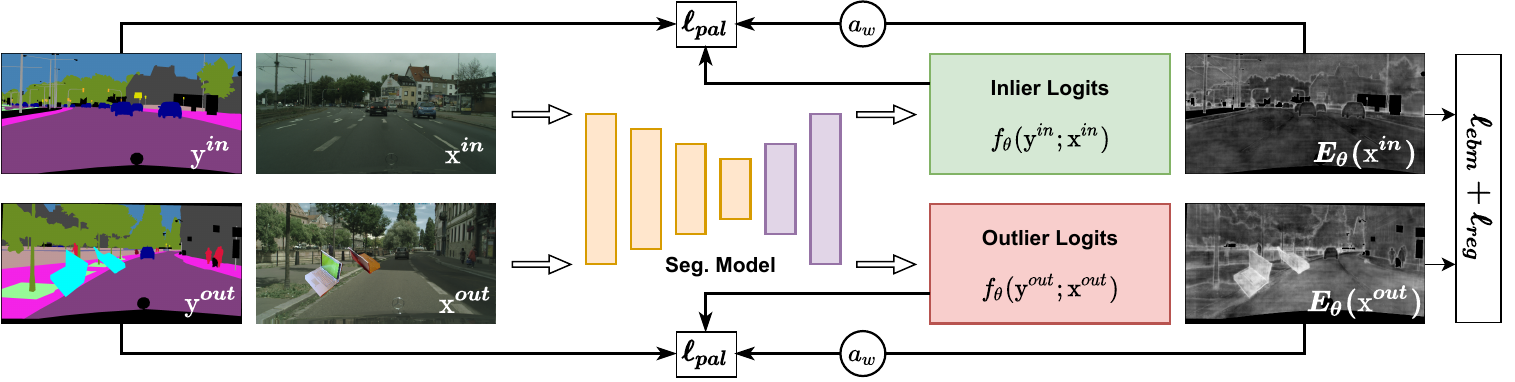}
    \vspace{-8pt}
    \caption{\textbf{PEBAL}. The pixel-wise anomaly abstention (PAL) loss $\ell_{pal}$ learns to abstain the prediction of outlier pixels from $\mathbf{x}^{out}$ containing OE objects (i.e., cyan coloured masks) and calibrate the logit of inlier classes (i.e., reduction of the inlier logits) from both inlier image $\mathbf{x}^{in}$ and outlier image $\mathbf{x}^{out}$. The EBM loss $\ell_{ebm}$ pushes the free energy $E_{\theta}$ to low values for inlier pixels and pulls that to high values for outlier pixels, where a regularisation loss $\ell_{reg}$ enforces the smoothness and sparsity constraints on the energy maps. Such EBM learning reduces the logit of inlier classes to share similar values at the same time, facilitating the $\ell_{pal}$ learning. 
    Then, the pixel-wise penalty $a_{\omega}$ associated with the abstention class at position $\omega$ is estimated to bias the penalty to be low for outlier pixels and high for inlier pixels, which in turn encourages
    high free energy for anomalies and enforces $\ell_{pal}$ to abstain the anomalous pixels.}
    \label{fig:model}
    \vspace{-10pt}
\end{figure*}

We present our PEBAL in this section (see Fig.~\ref{fig:model}), where we first describe the dataset, then  introduce abstention learning and EBM. Next, we present the loss function to train the model, followed by the training and inference procedures. 

\vspace{-10pt}
\subsection{Training Set}

We assume to have a set of inlier training images and annotations $\mathcal{D}^{in} = \{ (\mathbf{x}_i,\mathbf{y}^{in}_i) \}_{i=1}^{|\mathcal{D}^{in}|}$, where
$\mathbf{x} \in \mathcal{X} \subset \mathbb{R}^{H \times W \times C}$ denotes an image with $C$ colour channels, and
$\mathbf{y}^{in} \in \mathcal{Y}^{in} \subset \{0,1\}^{H \times W \times Y}$ denotes the inlier pixel level labels that can belong to $Y$ classes. 
We also have a set of outlier images and annotations $\mathcal{D}^{out} = \{ (\mathbf{x}_i,\mathbf{y}^{out}_i) \}_{i=1}^{|\mathcal{D}^{out}|}$, where
$\mathbf{y}^{out} \in \mathcal{Y}^{out} \subset \{0,1\}^{H \times W \times (Y+1)}$ 
denotes the outlier pixel-level labels, with the class $Y+1$ reserved for pixels belonging to the anomaly class. 
Note that similarly to previous papers~\cite{chan2021entropy}, the types of anomalies in training set $\mathcal{D}^{out}$ do not overlap with the anomalies to be found in the testing set.




\vspace{-10pt}
\subsection{Pixel-wise Energy-biased Abstention Learning (PEBAL)}

The PEBAL model is denoted by
\begin{equation}
    p_{\theta}(y|\mathbf{x})_{\omega}=\frac{\exp(f_{\theta}(y ; \mathbf{x})_{\omega})}{\sum_{y' \in \{1,...,Y+1\}}\exp(f_{\theta}(y' ; \mathbf{x})_{\omega})},
    \label{eq:pixel_wise_segmentation}
\end{equation}
where $\theta$ is the model parameter,
$\omega$ indexes a pixel in the image lattice $\Omega$, $p_{\theta}(y|\mathbf{x})_{\omega}$ represents the probability of labelling pixel $\omega$ with $y \in \{1,...,Y+1\}$, and $f_{\theta}(y ; \mathbf{x})_{\omega}$ is the logit for class $y$ at pixel $\omega$.

To train the model in~\eqref{eq:pixel_wise_segmentation}, we formulate a cost function that jointly trains PAL and EBM to classify anomalous pixels.
An important training hyper-parameter for PAL is the penalty to abstain from the classification into one of the inlier classes in $\{1,...,Y\}$--this penalty is generally tuned to a single value for all training samples through model selection (e.g., cross validation)~\cite{liu2019deep}. 
Instead of treating this as a tunable hyper-parameter, we propose the use of EBM (defined below in~\eqref{eq:energy_sample}) to automatically estimate this penalty during the training process for each pixel within each training image.
More specifically, the cost function to train the PEBAL model in~\eqref{eq:pixel_wise_segmentation} is:
\begin{equation}
\begin{split} 
    \ell & (\mathcal{D}^{in},   \mathcal{D}^{out},\theta) = \\ & \sum_{(\mathbf{x},\mathbf{y}^{in})\in \mathcal{D}^{in}} \big ( \ell_{pal}(\theta,\mathbf{y}^{in},\mathbf{x},E_{\theta}(\mathbf{x})) + \lambda \ell^{in}_{ebm}(E_{\theta}(\mathbf{x}))  + \ell_{reg}(E_{\theta}(\mathbf{x})) \big ) + \\
    & \sum_{(\mathbf{x},\mathbf{y}^{out})\in \mathcal{D}^{out}} \big ( \ell_{pal}(\theta,\mathbf{y}^{out},\mathbf{x},E_{\theta}(\mathbf{x})) + \lambda \ell^{out}_{ebm}(E_{\theta}(\mathbf{x}))   + \ell_{reg}(E_{\theta}(\mathbf{x})) \big ).
\end{split}%
\label{eq:full_loss}
\end{equation}
where $\ell_{pal}(.)$ denotes the PAL loss defined as
\begin{equation}
    \ell_{pal}(\theta,\mathbf{y},\mathbf{x},E_{\theta}(\mathbf{x})) = -\sum_{\omega \in \Omega} \log \Big ( f_{\theta}(y_{\omega};\mathbf{x})_{\omega} + \frac{f_{\theta}(Y+1;\mathbf{x})_{\omega}}{a_{\omega}} \Big ),%
\label{eq:loss_AL}
\end{equation}
with $y_{\omega} \in \{1,...,Y\}$ for $\mathbf{y}^{in}$, $y_{\omega} \in \{1,...,Y+1\}$ for $\mathbf{y}^{out}$\footnote{When $y_{\omega}$ is an outlier pixel, we set $y_{\omega}=1$ for all the $Y+1$ labels in $\mathcal{Y}^{out}$.},
and $a_{\omega}$ denotes the pixel-wise penalty associated with abstaining from the classification of the inlier classes.
The minimisation of the loss in~\eqref{eq:loss_AL} will abstain from classifying outlier pixels into one of the inlier classes, where a pixel is estimated to be an outlier with $a_{\omega}$. 
Before formulating $a_{\omega}$, let us define 
the inlier free energy at pixel $\omega$, which is denoted by $E_{\theta}(\mathbf{x})_{\omega}$ and computed with the $logsumexp$ operator as follows~\cite{lecun2006tutorial,liu2020energy,grathwohl2019your}:
\begin{equation}
E_{\theta}(\mathbf{x})_{\omega}=-\log\sum_{y \in \{1,...,Y\}}\exp(f_{\theta}(y;\mathbf{x})_{\omega}).
\label{eq:energy_sample}
\end{equation}
The pixel-wise penalty associated with abstaining from the classification of the inlier classes is defined by
\begin{equation}
    a_{\omega}=(-E_{\theta}(\mathbf{x})_{\omega})^2,
    \label{eq:def_a_w}
\end{equation}
which means that the larger the $a_{\omega}$ (i.e., low inlier free energy, so the sample is an inlier), the higher the loss to abstain from classifying into one of the $Y$ classes, and low value of $a_{\omega}$ (i.e., high free inlier energy, which means an outlier sample) implies a lower loss to abstain from classifying one of the $Y$ classes.
Also in~\eqref{eq:full_loss},
$\ell^{in}_{ebm}(.)$ (weighted by hyper-parameter $\lambda$)
represents the EBM loss that pushes the inlier free energy in~\eqref{eq:energy_sample} for samples in $\mathcal{D}^{in}$ to low values, with
\begin{equation}
    \ell_{ebm}^{in}(E_{\theta}(\mathbf{x}))=\sum_{\omega \in \Omega} \big ( \max(0,E_{\theta}(\mathbf{x})_{\omega}-m_{in}) \big )^2,
    \label{eq:loss_ebm_in}
\end{equation}
representing the loss of having inlier samples with free energy larger than threshold $m_{in}$, and 
\begin{equation}
    \ell_{ebm}^{out}(E_{\theta}(\mathbf{x}))=\sum_{\omega_{in} \in \Omega} \big ( \max(0,E_{\theta}(\mathbf{x})_{\omega}-m_{in}) \big )^2 +  \sum_{\omega_{out} \in \Omega} \big ( \max(0,m_{out}-E_{\theta}(\mathbf{x})_{\omega}) \big )^2,
    \label{eq:loss_ebm_out}
\end{equation}
denoting the loss of having outlier samples with inlier free energy smaller than threshold $m_{out}$, where the margin losses in~\eqref{eq:loss_ebm_in} 
and~\eqref{eq:loss_ebm_out} effectively create an energy gap between normal and abnormal pixels\footnote{Please note that $\mathcal{D}^{out}$ contains both iniler and outlier pixels. }.
The last term to define in~\eqref{eq:full_loss} is the inlier free energy regularisation loss to enforce that anomalous pixels are sparse and pixel anomaly classification is smooth (i.e., anomalous pixels tend to have anomalous neighbouring pixels), which is defined as
\begin{equation}
    \ell_{reg}(E_{\theta}(\mathbf{x})) = \sum_{\omega \in \Omega} 
    \beta_1 | E_{\theta}(\mathbf{x})_{\omega} - E_{\theta}(\mathbf{x})_{\mathcal{N}(\omega)} | + 
    \beta_2 | E_{\theta}(\mathbf{x})_{\omega} |,%
    \label{eq:loss_regularisation}
\end{equation}
where $\beta_1$ and $\beta_2$ are hyper-parameters that weight the contributions of the smoothness and sparsity and sparsity regularisations, and $\mathcal{N}(\omega)$  denotes neighbouring pixels in horizontal and vertical directions.

\vspace{-10pt}
\subsection{Training and Inference}
\label{sec:train_inference}

\noindent\textbf{Training.} An important point of the training process is how to setup the inlier and outlier datasets $\mathcal{D}^{in}$ and $\mathcal{D}^{out}$.
A recently published paper~\cite{chan2021entropy} carefully selects images to be included in $\mathcal{D}^{out}$ by making sure that 
the segmentation labels presented in those images do not overlaps with the inlier labels. In particular for~\cite{chan2021entropy},  $\mathcal{D}^{in}$ has images and annotations from Cityscape and $\mathcal{D}^{out}$ has images and annotations from COCO~\cite{lin2014microsoft}. We argue that there are two issues with this strategy to form $\mathcal{D}^{out}$, which are: 1) the selected COCO images generally only contain anomalous pixel labels, leading to unstable training 
of the outlier losses (i.e., second summation in~\eqref{eq:full_loss}) given the exclusive presence of the anomaly class (in effect, this becomes a one-class segmentation problem);  
2) re-training the model with images containing only anomalous pixels removes the semantic context of inlier pixels when training for the outlier losses, which 
can deteriorate the segmentation accuracy of the inlier labels. 

To mitigate these issues, 
we form $\mathcal{D}^{out}$ using a novel extension based on CutMix and CutPaste~\cite{yun2019cutmix,li2021cutpaste}, which we refer to as AnomalyMix.
AnomalyMix cuts the anomalous objects
from an outlier dataset (e.g., COCO) using its labelled masks and paste them into the images of the inlier dataset (e.g., CitySpace), where we label the pixels of the anomalous object with the class $Y+1$ -- these images are then inserted into $\mathcal{D}^{out}$.
AnomalyMix addresses the two issues above because the outlier images now contain a combination of inlier and outlier pixels, allowing a balanced learning and keeping the visual context of inlier labels when training for the outlier losses.
Furthermore, AnomalyMix can form a potentially infinite number of training images for $\mathcal{D}^{out}$ given the range of transformations to be applied to the cut objects and the locations of the inlier images that the objects can be pasted.
Previous papers~\cite{di2021pixel,jung2021standardized} argue that re-training the whole segmentation model can jeopardise the segmentation accuracy for the inlier classes. Furthermore, such re-training requires a long training time, leading to inefficient optimisation. 
In this work, we propose to \textbf{fine-tune only the final classification block} 
using the loss in~\eqref{eq:full_loss}, instead of re-training the whole segmentation model.
Besides being efficient, this fast fine-tuning keeps the segmentation accuracy of the model in the original dataset used for pre-training the model. 
Furthermore, an interesting side-effect of our training is that the cost function in~\eqref{eq:full_loss} will calibrate the segmentation prediction for the inlier classes. 
This happens because the terms $\ell_{pal}(.)$,  $\ell^{in}_{ebm}(.)$ and $\ell^{out}_{ebm}(.)$ jointly constrain the maximisation of logits and naturally calibrate classification confidence (See supplementary material). 

\vspace{3pt}
\noindent \textbf{Inference.} During inference, pixel-wise anomaly detection is performed by computing the inlier free energy score $\mathbb{E}_{\theta}(\mathbf{x})_{\omega}$ from~\eqref{eq:energy_sample} for each pixel position $\omega$ given a test image $\mathbf{x}$ and inlier segmentation is obtained from the inlier classes from the PEBAL model in~\eqref{eq:pixel_wise_segmentation}. Following~\cite{jung2021standardized}, we also apply a Gaussian smoothing kernel to produce the final energy map.

\vspace{-10pt}
\section{Experiment}
\label{sec:experiment}

\vspace{-5pt}
\subsection{Datasets}
\noindent \textbf{LostAndFound}~\cite{pinggera2016lost} is one of the first publicly available urban driving scene anomaly detection datasets  containing real-world anomalous objects. The dataset has an official testing set containing 1,203 images with small obstacles in front of the cars, collecting from 13 different street scenes, featuring 37 different types of anomalous objects with various sizes and material. 

\noindent \textbf{Fishyscapes}~\cite{blum2019fishyscapes} is a high-resolution dataset for anomaly estimation in semantic segmentation for urban driving scenes. The benchmark has an online testing set that is entirely unknown to the methods. The dataset is composed by two data sources: Fishyscapes LostAndFound that contains a set of real road anomalous objects~\cite{pinggera2016lost} 
and a blending-based Fishyscapes Static dataset. 
The Fishyscapes LostAndFound validation set consists of 100 images from the aforementioned
LostAndFound dataset with refined labels and the Fishyscapes Static validation set contains 30 images with the blended anomalous objects from Pascal VOC~\cite{everingham2010pascal}. 
For all datasets, we select the checkpoints based on the results on the public validation sets, but submitted our code and checkpoints to the benchmark website to be evaluated on their hidden test sets. 

\noindent \textbf{Road Anomaly}~\cite{lis2019detecting} contains real-world road anomalies in front of the vehicles. The dataset has 60 images from the Internet, containing unexpected animals rocks, cones and obstacles.Unlike the LostAndFound and Fishyscapes, this dataset contains abnormal objects with various scales and sizes, making it even more challenging.  
\vspace{-10pt}
\subsection{Implementation Details}

Following~\cite{chan2021entropy,chan2021segmentmeifyoucan}, we use DeepLabv3+~\cite{chen2018encoder} with WideResnet38 trained by Nvidia~\cite{zhu2019improving} and ResNet101 from~\cite{jung2021standardized} as the backbone of our segmentation models. The training details of those models can be found in their original papers or our supplementary material.
The models are trained on Cityscapes~\cite{cordts2016cityscapes} training set. For our PEBAL fine-tuning, we empirically set the $m_{in}$ and $m_{out}$ in Eq.~\ref{eq:loss_ebm_in} and Eq.~\ref{eq:loss_ebm_out} as -12 and -6, respectively. The weights $\beta_{1}$ and $\beta_{2}$ in Eq.~\ref{eq:loss_regularisation} are set to $5e-4$ and $3e-6$~\cite{sultani2018real}, $\lambda$ in Eq.~\ref{eq:full_loss} to 0.1, and the weight of $\ell_{ebm}$ to 0.1, respectively. Note that those hyper-parameters are selected at the first training epoch to normalise loss values to a similar scale. We also show our model can obtain consistently SOTA results regardless of the selection of hyper-parameters in the supplementary material. 
Our training consists of fine-tuning the final classification block of the model for 20 epochs. We use the same resolution of random crop as in~\cite{zhu2019improving}, and use Adam with a learning rate of $1e^{-5}$. The batch size is set to 16. 
Following~\cite{chan2021entropy}, for our AnomalyMix augmentation, we randomly sample 297 images as training data from the remaining COCO images that do not contain objects in Cityscapes or our anomaly validation/testing sets and randomly apply AnomalyMix to mix them into the Cityscape training images, following Chan et al.~\cite{chan2021entropy}.


\vspace{-10pt}
\subsection{Evaluation Measures}

Following~\cite{jung2021standardized,blum2019fishyscapes,di2021pixel,chan2021entropy}, we compute the the area under receiver operating characteristics (AUROC), average precision (AP), and the false positive rate at a true positive rate of 95\% (FPR95) to validate our approach. 
For Fishyscapes public leaderboard, we use AP and FPR95 to compare with other methods, same as their website.


\vspace{-10pt}
\subsection{Comparison on Anomaly Segmentation Benchmarks}

\begin{table}[t!]
\centering
\caption{Anomaly segmentation results on \textbf{LostAndFound} testing set, with \textbf{WideResnet38} backbone. All methods use the same segmentation models. * indicate that the model requires additional learnable parameters. $\dagger$ indicates that the results are obtained from the official code with our WideResnet38 backbone. 
}
\resizebox{0.42\linewidth}{!}{%
\begin{tabular}{@{}cccc@{}}
\toprule
Methods & AUC $\uparrow$ & AP $\uparrow$ & FPR$_{95}$ $\downarrow$ \\ \hline \hline
MSP~\cite{hendrycks2019scaling} & 85.49 & 38.20 & 18.56  \\
Mahalanobis~\cite{lee2018simple} & 79.53  & 42.56  & 24.51  \\
Max Logit~\cite{hendrycks2016baseline} & 94.52  & 65.45  & 15.56  \\
Entropy~\cite{hendrycks2016baseline} & 86.52  & 50.66  & 16.95  \\
Energy~\cite{liu2020energy} & 94.45  & 66.37  & 15.69    \\ 
Meta-OoD~\cite{chan2021entropy} & 97.95  &  71.23   & 5.95   \\
$^{\dagger}$SML~\cite{jung2021standardized} & 88.05 & 25.89  & 44.48    \\
$^{\dagger}$SynBoost*~\cite{di2021pixel} & 98.38 & 70.43  & 4.89    \\
Deep Gambler~\cite{liu2019deep} & 98.67   & 72.73  & 3.81    \\\midrule
Ours & \textbf{99.76} & \textbf{78.29} & \textbf{0.81}  \\ \bottomrule
\end{tabular}%
}
\label{tab:lost_found_test}
\vspace{-10pt}
\end{table}

\vspace{-5pt}
\subsubsection{Comparison on LostAndFound.}
Table~\ref{tab:lost_found_test} shows the result on the testing set of LostAndFound. Notably, our approach surpasses the previous baseline approaches (i.e., MSP~\cite{hendrycks2019scaling}, Mahalanobis~\cite{lee2018simple}, Max Logit~\cite{hendrycks2016baseline} and Entropy~\cite{hendrycks2016baseline}) by 10\% to 40\% AP, and 13\% to 22\% FPR95, respectively. 
When compared with previous SOTA approaches such as SynBoost~\cite{di2021pixel}, SML~\cite{jung2021standardized} and Meta-OoD~\cite{chan2021entropy}, we improve the AP performance by a large margin (15\% to 40\%), and decrease the FPR95 by about 5\% to 70\%. This illustrates the robustness and effectiveness on detecting small and distant anomalous objects given that the dataset contains mostly real-world small objects. 
Our PEBAL also improves the EBM baseline~\cite{liu2020energy} and the AL baseline based on Deep Gambler~\cite{liu2019deep}. This demonstrates that a simple adaptation of AL and EBM is not enough to enable accurate pixel-wise anomaly detection. Previous SOTA SML~\cite{jung2021standardized} aims to balance the inlier class-wise discrepancy on prediction scores, which is disadvantageous for measuring performance on LostAndFound test set since there may be no classes in the evaluation other than the road class (i.e., most of the inlier classes within LF test set is road class), thus leading to significant performance variations between LostAndFound and Fishyscapes. 
It is worth noting that our approach achieves 1.03\% FPR95, significantly reducing the false positive pixels, improving the chances of applying it to real-world applications. 
%

\vspace{-10pt}
\subsubsection{Comparison on Fishyscapes Leaderboard.}
Table~\ref{tab:public_website} shows the leaderboard results on the test set of Fishyscapes LostAndFound and Fishyscapes Static. Following~\cite{jung2021standardized}, we compared the methods based on whether they require re-training of the entire segmentation network, adding the extra network, or utilising the OoD data. We achieve the SOTA performance by a large margin on Fishyscapes leaderboard when compared with the previous methods except \cite{bevandic2019simultaneous} (Static) that rely on an inefficient re-training segmentation model, extra learnable parameters, and extra OoD training data. Without re-training the entire network or adding extra learnable parameters, our approach can work efficiently to surpass previous SOTA competing approaches that fall into the same category by about 13\% to 42\% on LostAndFound and 40\% to 50\% AP on Static. 
Such significant improvements indicate the generalisation ability of our proposed PEPAL on detecting a wide variety of unseen abnormalities (i.e., of different size, type, scene, and distance) substantially reducing false negative and positive pixels. 
Moreover, it is worth noting that  PEBAL reduces the amount of false positive pixels to 7.58 and 1.73 FPR on the two datasets. This result is publicly available on the Fishyscapes website. 

\begin{table*}[!t]
\centering
\caption{Comparison with previous approaches on \textbf{Fishyscapes Leaderboard}. We achieve a new state-of-the-art performance among the approaches that require extra OoD data, and without re-training the segmentation networks and extra networks on Fishyscapes Leaderboard.
}
\resizebox{0.8\linewidth}{!}{%
\begin{tabular}{c|c|c|c|cc|cc}
\toprule
\multirow{2}{*}{Models} & \multirow{2}{*}{re-training} & \multirow{2}{*}{Extra Network} & \multirow{2}{*}{OoD Data} & \multicolumn{2}{c|}{FS LostAndFound} & \multicolumn{2}{c}{FS Static}  \\ \cline{5-8}
 &  &  &  & AP $\uparrow$ & FPR95 $\downarrow$ & AP $\uparrow$ & FPR95 $\downarrow$  \\ \hline \hline
Discriminative Outlier Detection Head~\cite{bevandic2019simultaneous}   &
  \multicolumn{1}{c|}{\textcolor{green}{\CheckmarkBold}} &
  \textcolor{green}{\CheckmarkBold} &
  \textcolor{green}{\CheckmarkBold} &
  31.31 &
  19.02 &
  96.76 &
  0.29  \\\hline
  MSP~\cite{hendrycks2019scaling}                             & \multicolumn{1}{c|}{\textcolor{red}{\XSolidBold}}    & \textcolor{red}{\XSolidBold}    & \textcolor{red}{\XSolidBold}    & 1.77  & 44.85 & 12.88 & 39.83    \\
Entropy~\cite{hendrycks2016baseline}                         & \multicolumn{1}{c|}{\textcolor{red}{\XSolidBold}}    & \textcolor{red}{\XSolidBold}    & \textcolor{red}{\XSolidBold}    & 2.93  & 44.83 & 15.41 & 39.75    \\
SML~\cite{jung2021standardized}                           & \multicolumn{1}{c|}{\textcolor{red}{\XSolidBold}}    & \textcolor{red}{\XSolidBold}    & \textcolor{red}{\XSolidBold}    & 31.05 & 21.52 & 53.11 & 19.64   \\ kNN Embedding - density~\cite{blum2019fishyscapes}         & \multicolumn{1}{c|}{\textcolor{red}{\XSolidBold}}    & \textcolor{red}{\XSolidBold}    & \textcolor{red}{\XSolidBold}    & 3.55  & 30.02 & 44.03 & 20.25    \\
Bayesian Deeplab~\cite{mukhoti2018evaluating}                & \multicolumn{1}{c|}{\textcolor{green}{\CheckmarkBold}} & \textcolor{red}{\XSolidBold}    & \textcolor{red}{\XSolidBold}    & 9.81  & 38.46 & 48.70  & 15.05   \\
Density - Single-layer NLL~\cite{blum2019fishyscapes}      & \multicolumn{1}{c|}{\textcolor{red}{\XSolidBold}}    & \textcolor{green}{\CheckmarkBold} & \textcolor{red}{\XSolidBold}    & 3.01  & 32.9  & 40.86 & 21.29   \\
Density - Minimum NLL~\cite{blum2019fishyscapes}           & \multicolumn{1}{c|}{\textcolor{red}{\XSolidBold}}    & \textcolor{green}{\CheckmarkBold} & \textcolor{red}{\XSolidBold}    & 4.25  & 47.15 & 62.14 & 17.43  \\
Image Resynthesis~\cite{lis2019detecting}                & \multicolumn{1}{c|}{\textcolor{red}{\XSolidBold}}    & \textcolor{green}{\CheckmarkBold} & \textcolor{red}{\XSolidBold}    & 5.70   & 48.05 & 29.6  & 27.13  \\
OoD Training - Void Class     & \multicolumn{1}{c|}{\textcolor{green}{\CheckmarkBold}} & \textcolor{red}{\XSolidBold}    & \textcolor{green}{\CheckmarkBold} & 10.29 & 22.11 & 45.00    & 19.40     \\
  
Dirichlet Deeplab~\cite{malinin2018predictive}              & \multicolumn{1}{c|}{\textcolor{green}{\CheckmarkBold}} & \textcolor{red}{\XSolidBold}    & \textcolor{green}{\CheckmarkBold} & 34.28 & 47.43 & 31.30  & 84.60    \\
Density - Logistic Regression~\cite{blum2019fishyscapes}   & \multicolumn{1}{c|}{\textcolor{red}{\XSolidBold}}    & \textcolor{green}{\CheckmarkBold} & \textcolor{green}{\CheckmarkBold} & 4.65  & 24.36 & 57.16 & 13.39   \\
SynBoost~\cite{di2021pixel}                      & \multicolumn{1}{c|}{\textcolor{red}{\XSolidBold}}    & \textcolor{green}{\CheckmarkBold} & \textcolor{green}{\CheckmarkBold} & 43.22 & 15.79 & 72.59 & 18.75        \\ \hline
Ours                          & \multicolumn{1}{c|}{\textcolor{red}{\XSolidBold}} & \textcolor{red}{\XSolidBold}    & \textcolor{green}{\CheckmarkBold} & \textbf{44.17}    &     \textbf{7.58}   &   \textbf{92.38}     &  \textbf{1.73}                 \\ \bottomrule
\end{tabular}%
}
\label{tab:public_website}
\vspace{-10pt}
\end{table*}

\begin{table*}[!t]
\centering
\caption{Anomaly segmentation results on \textbf{Fishyscapes validation sets} (LostAndFound and Static), and the \textbf{Road Anomaly testing set}, with \textbf{WideResnet38} backbone. * indicate that the model requires additional learnable parameters. $\dagger$ indicates that the results are obtained from the official code with our WideResnet38 backbone.  
Best and second best results in bold. 
}
\resizebox{.8\linewidth}{!}{$%
\begin{tabular}{@{}c|ccc|ccc|ccc@{}}
\toprule
\multirow{2}{*}{Methods} & \multicolumn{3}{c|}{FS LostAndFound} & \multicolumn{3}{c|}{FS Static} & \multicolumn{3}{c}{Road Anomaly}  \\ \cline{2-10}
 & AUC $\uparrow$  & AP $\uparrow$  & FPR$_{95}$ $\downarrow$  & AUC $\uparrow$  & AP $\uparrow$  & FPR$_{95}$ $\downarrow$ & AUC $\uparrow$  & AP $\uparrow$  & FPR$_{95}$ $\downarrow$  \\ \hline  \hline
MSP~\cite{hendrycks2019scaling} & 89.29  & 4.59  & 40.59  & 92.36  & 19.09  & 23.99  & 67.53  & 15.72  & 71.38   \\
Max Logit~\cite{hendrycks2019scaling} & 93.41  & 14.59  & 42.21  & 95.66  & 38.64  & 18.26  & 72.78  & 18.98  & 70.48   \\
Entropy~\cite{hendrycks2016baseline} & 90.82  & 10.36  & 40.34  & 93.14  & 26.77  & 23.31  & 68.80  & 16.97  & 71.10   \\
Energy~\cite{liu2020energy} & 93.72  & 16.05  & 41.78  & 95.90  & 41.68  & 17.78  & 73.35  & 19.54  & 70.17  \\ 
Mahalanobis~\cite{lee2018simple} & 96.75  & 56.57 & 11.24  & 96.76  & 27.37  & 11.7  & 62.85  & 14.37  & 81.09  \\
Meta-OoD~\cite{di2021pixel} & 93.06  & 41.31  & 37.69  & 97.56  & 72.91  & 13.57  & - & - & -  \\
$^{\dagger}$Synboost*~\cite{di2021pixel} & 96.21 & \textbf{60.58} & 31.02 & 95.87  & 66.44  & 25.59 & \textbf{81.91}  & \textbf{38.21} & \textbf{64.75}    \\
$^{\dagger}$SML~\cite{jung2021standardized} & 94.97	& 22.74 &	33.49 &	97.25 &	66.72 &	12.14 &	75.16 &	17.52 &	70.70    \\
Deep Gambler~\cite{liu2019deep} & \textbf{97.82}  & 31.34  & \textbf{10.16}  & \textbf{98.88}  &  \textbf{84.57}  & \textbf{3.39}  & 78.29 & 23.26 & 65.12   \\ \midrule
Ours & \textbf{98.96}  & \textbf{58.81}  & \textbf{4.76}  & \textbf{99.61}  & \textbf{92.08}  & \textbf{1.52}  & \textbf{87.63}  & \textbf{45.10}  & \textbf{44.58}  \\ \bottomrule
\end{tabular}%
$} 
\label{tab:wres34_fishy}
\vspace{-10pt}
\end{table*}

\begin{table*}[t!]
\centering
\caption{Anomaly segmentation results on \textbf{Fishyscapes validation sets} (LostAndFound and Static), and the \textbf{Road Anomaly testing set}, with \textbf{Resnet101} backbone. * indicate that the model requires additional learnable parameters. $\dagger$ indicates that the results are obtained from the official code with our Resnet101 backbone. Best and second best results in bold.}
\resizebox{0.86\linewidth}{!}{%
\begin{tabular}{@{}c|ccc|ccc|ccc@{}}
\toprule
\multirow{2}{*}{Methods} & \multicolumn{3}{c|}{FS LostAndFound} & \multicolumn{3}{c|}{FS Static} & \multicolumn{3}{c}{Road Anomaly}  \\ \cline{2-10}
 & AUC $\uparrow$  & AP $\uparrow$  & FPR$_{95}$ $\downarrow$  & AUC $\uparrow$  & AP $\uparrow$  & FPR$_{95}$ $\downarrow$ & AUC $\uparrow$  & AP $\uparrow$  & FPR$_{95}$ $\downarrow$  \\ \hline  \hline
MSP~\cite{hendrycks2019scaling} & 86.99 & 6.02 & 45.63 & 88.94 & 14.24 & 34.10 & 73.76 & 20.59 & 68.44  \\
Max Logit~\cite{hendrycks2019scaling} & 92.00 & 18.77 & 38.13 & 92.80 & 27.99 & 28.50 & 77.97 & 24.44 & 64.85  \\
Entropy~\cite{hendrycks2016baseline}  & 88.32 & 13.91 & 44.85 & 89.99 & 21.78 & 33.74 & 75.12 & 22.38 & 68.15  \\
Energy~\cite{liu2020energy}  & 93.50  & 25.79 & 32.26 & 91.28	& 31.66	& 37.32	& 78.13	& 24.44	& 63.36  \\ 
$^{\dagger}$SynthCP*~\cite{xia2020synthesize} & 88.34 & 6.54 & 45.95 & 89.9 & 23.22 & 34.02 & 76.08 & 24.86 & 64.69  \\
$^{\dagger}$Synboost*~\cite{di2021pixel} & 94.89  & \textbf{40.99} & 34.47  & 92.03  & 48.44 & 47.71  & 85.23 & \textbf{41.83} & 59.72  \\
SML~\cite{jung2021standardized} & 96.88 & 36.55 & 14.53 & 96.69 & 48.67 & 16.75 & 81.96 & 25.82 & 49.74  \\
Deep Gambler~\cite{liu2019deep} & \textbf{97.19} & 39.77   & \textbf{12.41}   & \textbf{97.51}  & \textbf{67.69}   & \textbf{15.39}  &  \textbf{85.45}  & 31.45  & \textbf{48.79}    \\
\midrule
Ours & \textbf{99.09} & \textbf{59.83} & \textbf{6.49} & \textbf{99.23} & \textbf{82.73} & \textbf{6.81} & \textbf{92.51} & \textbf{62.37} & \textbf{28.29}  \\ \bottomrule
\end{tabular}%
}
\label{tab:res101_val}
\vspace{-10pt}
\end{table*}

\vspace{-10pt}
\subsubsection{Comparison on Fishyscapes validation sets and Road Anomaly.}
In Tables~\ref{tab:wres34_fishy} and \ref{tab:res101_val}, we compare our approach on the Fishyscapes validation sets and Road Anomaly using two different backbones. Our model outperforms the previous methods by a large margin on all three benchmarks, regardless of the backbones and their segmentation accuracy. 
To verify the applicability of our method, except for the modern WideResnet38 backbone,
we use a ResNet101 DeepLabv3+ 
to investigate the performance in terms of the size of the architecture and its inlier segmentation accuracy. The results demonstrate that our approach is applicable to a wide-range of segmentation models,
indicating the effectiveness of PEBAL to adapt to real-world systems. 

Moreover, our fine-tuning sacrifices only marginally the inlier segmentation accuracy (i.e., 0.2\% - 0.7\% mIoU on Cityscapes) for both backbones, achieving good performance on both inlier and anomaly segmentation. 
We present details of all inlier segmentation models (i.e., Cityscapes training setup and mIoU), and include more experimental results of other DeepLabv3+ checkpoints in supplementary material.

\vspace{-10pt}
\subsubsection{Remarks -- Superior Performance on Challenging Benchmarks.}
Each dataset has different challenges. For example, the LostAndFound testing set considers only drivable areas with homogeneous normal scenes  (i.e., road) and limited categories of abnormalities (i.e., road obstacles), leading to a relatively less challenging benchmark on which most methods can obtain  good AUC performance, as shown in Tables \ref{tab:lost_found_test}, \ref{tab:wres34_fishy} and \ref{tab:res101_val}. 
On the contrary, Fishyscapes and RoadAnomaly contain large number of heterogeneous inlier and outlier pixels from diverse classes, leading to significantly more difficult testbeds than the LostAndFound testing set. 
Furthermore,  Fishyscapes and RoadAnomaly contain domain shift compared with Cityscapes (e.g., both datasets contain different scenes than Cityscapes)
and have different types/sizes of OoD objects.
Most existing SOTA methods work ineffectively on these two datasets due to those challenges, while our adaptive pixel-level anomaly class learning helps our model effectively detect these challenging inlier and outlier pixels in the aforementioned heterogeneous and domain-shifted scenes, 
yielding substantial improvements (i.e., 20\% to 50\%) to previous approaches, as shown in Tables \ref{tab:public_website}, \ref{tab:wres34_fishy} and \ref{tab:res101_val}.

\vspace{-10pt}
\subsection{Ablation Study}
Table~\ref{tab:ablation} shows the contribution of each component of our PEBAL on the LostAndFound testing set. All modules are trained with COCO OE images using AnomalyMix.   
Adding an extra OoD class to learn the OE training samples with entropy maximisation (\textbf{EM}) is our baseline (first row). To justify the effectiveness of our proposed joint training, we show the results using 
energy-based models ($\ell_{ebm}$ without $\ell_{pal}$) and pixel-wise abstention ($\ell_{pal}$ with pre-defined fixed penalty). Both outperform the baselines (AP=70.2, FPR=8.9 and AP=72.7, FPR=3.8 vs. AP=69, FPR=8.03), while our proposed joint training ($\ell_{ebm}$ + $\ell_{pal}$) obtains 77.19\% of AP and 1.19\% of FPR, improving over each module by 4\% to 7\%. This indicates the effectiveness of our joint training and the significance of our proposed PAL with learnable adaptive energy-based penalties $a_{\omega}$. Finally, the smoothness and sparsity regularisation losses stabilise the training and further improve the performance. 

\begin{table}[t!]
\centering
\caption{Ablation studies for anomaly segmentation on \textbf{LostAndFound}, with \textbf{WideResnet38} backbone, where all proposed modules are trained with COCO OE images with AnomalyMix. EM denotes the baseline method that adds an extra OoD class to learn the OE training samples with entropy maximisation (first row).  }
\resizebox{.5\linewidth}{!}{%
\begin{tabular}{@{}c@{\hskip .15in}c@{\hskip .15in}c@{\hskip .15in}c@{\hskip .15in}c@{\hskip .15in}c@{\hskip .15in}c@{\hskip .15in}c@{\hskip .15in}c@{}}
\toprule
EM &  $\ell_{ebm}$ & $\ell_{pal}$ & $\ell_{reg}$ & AUC $\uparrow$ & AP $\uparrow$ & FPR$_{95}$ $\downarrow$\\ \hline \hline
 \checkmark &    & &  & 96.88  & 69.02  & 8.03  \\\hline
 &    \checkmark  &  &    & 97.88  & 70.24  & 8.92 \\ 
  &  & \checkmark  &  & 98.67   & 72.73 & 3.81 \\
 &    \checkmark & \checkmark & & 99.63  & 77.19  & 1.19\\ 
 &   \checkmark & \checkmark & \checkmark & \textbf{99.76} & \textbf{78.29} & \textbf{0.81}  \\
  \bottomrule
\end{tabular}%
}
\label{tab:ablation}
\vspace{-20pt}
\end{table}

\begin{table}[!t]
\centering
\caption{The performance comparison of our approach on Fishyscapes benchmark w.r.t different \textbf{diversity of OE classes} (mean results over six random seeds), in terms of AP and FPR95. } 
\resizebox{0.65\linewidth}{!}{
\begin{tabular}{!{\vrule width 1pt}c!{\vrule width 1pt}c|c!{\vrule width 1pt}c|c!{\vrule width 1pt}} 
\specialrule{1pt}{0pt}{0pt}
\multirow{2}{*}{Class Per.} & \multicolumn{2}{c!{\vrule width 1pt}}{FS LostAndFound }                    & \multicolumn{2}{c!{\vrule width 1pt}}{FS Static}  \\ 
\cline{2-5}
                                 & AP $\uparrow$                                  & FPR$_{95}$ $\downarrow$           & AP  $\uparrow$           & FPR$_{95}$ $\downarrow$              \\ 
\hline
1\%                           & 53.57 $\pm$3.74 & 6.97 {$\pm$1.98}  & 85.84 {$\pm$1.01} & 3.05 {$\pm$0.97}      \\ 
\hline
5\%                              & 52.16 $\pm$3.88 & 6.58 {$\pm$1.95}  & 90.57 {$\pm$1.75} & 1.93 {$\pm$0.52}      \\ 
\hline
10\%                             & 55.14 {$\pm$3.02}       & 5.78 {$\pm$1.59} & 91.37 {$\pm$1.28} & 1.64 {$\pm$0.58}      \\ 
\hline
25\%                             & 55.48 {$\pm$3.32}  & 5.98 {$\pm$1.27}   & 91.28 {$\pm$1.94} & 1.77 {$\pm$0.18}      \\ 
\hline
50\%                             & 56.69 {$\pm$2.57}  & 5.32 {$\pm$1.16}   & 91.88 {$\pm$0.71} & 1.62 {$\pm$0.05}      \\
\hline
75\%                             & 57.86 {$\pm$2.83}  & 5.11 {$\pm$1.69}   & 91.85 {$\pm$0.56} & 1.63 {$\pm$0.09}      \\ 
\hline
\specialrule{1pt}{0pt}{0pt}
\end{tabular}
}

\label{tab:coco_class}
\vspace{-10pt}
\end{table}

\vspace{-10pt}
\subsection{Outlier Samples and Computational Efficiency}
\vspace{-5pt}
\subsubsection{Outlier Diversity and Efficiency.} In Table~\ref{tab:coco_class}, we randomly select 1\%, 5\%, 10\%, 25\% 50\%, and 75\% of COCO classes as the OE data during training and compute the mean results over six different random seeds.  
We achieve consistent AP and FPR performance regardless of the number of COCO classes used during the training on Fishyscapes. It is also worth noting that our approach can effectively learn the PEBAL model using \textbf{only one class} (1\% in Table~\ref{tab:coco_class}) of outlier data, which selects some of the irrelevant classes of COCO objects that are not possible to be found on road in real life (e.g., dining table, laptop, and clock). The results indicate that our model can consistently achieve SOTA performance on Fishyscapes without a careful selection of OE classes, demonstrating the robustness of our approach under diverse outlier classes. 
We also investigate the outlier sample efficiency of our model w.r.t smaller OE training sets with a fixed 100\% COCO classes (80 classes) on Fishyscapes in Table~\ref{tab:coco_amount}, and we achieve consistently good performance regardless the number of outlier training samples.  All those experiments show the applicability of our PEBAL to real-world autonomous driving systems.


\vspace{-10pt}
\subsubsection{Computational Efficiency.}
We compare the computational efficiency of our PEBAL with previous SOTA Meta-OoD~\cite{chan2021entropy} and Synboost~\cite{di2021pixel} in terms of the trainable parameters, training time and mean inference time per image, on an NVIDIA3090. 
As PEBAL requires the fine-tuning of the final classification block, it has only 1.3M parameters and each training epoch takes about 12 minutes, which is significantly less than the re-training approach Meta-OoD that has 137.1M parameters and each training epoch takes about 26 minutes, and the reconstruction based approach Synboost that takes about 33 minutes to train a epoch of its re-synthesis and dissimilarity networks with 157.3M parameters. 
Moreover, our method also has a much faster mean inference time of 0.55s compared to 0.85s of Meta-OoD and 1.95s of Synboost. Those results suggest the practicability of our model in real-world self-driving systems. 


\begin{table}[!t]
\centering
\caption{The performance comparison of our approach on Fishyscapes benchmark w.r.t different \textbf{amount of OE training samples} (mean results over six random seeds), in terms of AP and FPR95.} 
\resizebox{.65\linewidth}{!}{
\begin{tabular}{!{\vrule width 1pt}c!{\vrule width 1pt}c|c!{\vrule width 1pt}c|c!{\vrule width 1pt}} 
\specialrule{1pt}{0pt}{0pt}
\multirow{2}{*}{Train Size} & \multicolumn{2}{c!{\vrule width 1pt}}{FS LostAndFound}                    & \multicolumn{2}{c!{\vrule width 1pt}}{FS Static}  \\ 
\cline{2-5}
                                 & AP $\uparrow$                                  & FPR$_{95}$ $\downarrow$           & AP  $\uparrow$           & FPR$_{95}$ $\downarrow$              \\ 
\hline
5\%                             & 54.32  {$\pm$1.89}  & 5.77 {$\pm$2.38}  & 	89.11 {$\pm$1.52} & 2.23 {$\pm$0.65}      \\ 
\hline 
10\%                             & 56.28 {$\pm$1.05}       & 4.66  {$\pm$1.36} & 90.02 {$\pm$0.57} & 1.67 {$\pm$0.28}      \\  
\hline
25\%                             & 56.18 {$\pm$1.69}  & 4.81 {$\pm$1.44}   & 91.23 {$\pm$0.95} & 1.63 {$\pm$0.22}      \\ 
\hline 
50\%                             & 57.34 {$\pm$1.19}  & 4.75  {$\pm$1.32}   & 91.29 {$\pm$0.92} & 1.67 {$\pm$0.17}      \\ 
\hline
\specialrule{1pt}{0pt}{0pt}
\end{tabular}
}
\label{tab:coco_amount}
\vspace{-10pt}
\end{table}

\vspace{-10pt}
\section{Conclusions and Discussions}
We proposed a simple yet effective approach, named Pixel-wise Energy-biased Abstention Learning (PEBAL), to fine-tune the last block of a segmentation model to detect unexpected road anomalies. 
The approach introduces a non-trivial training that jointly optimises a novel pixel-wise abstention learning and an energy-based model to learn an adaptive pixel-wise anomaly class, in which a new pixel-wise energy-biased penalty estimation method is proposed to improve the precision and robustness to detect small and distant anomalous objects. 
The resulting model significantly reduces the
false positive and false negative detected anomalies, compared with previous SOTA methods. The results on four benchmarks demonstrate the accuracy and robustness of our approach to detect anomalous objects regardless of the amount or diversity of exposed training outliers. Despite the remarkable performance on most datasets, PEBAL is not as effective on the most challenging dataset, Road Anomaly, that contains significantly more diverse and realistic anomalous objects. We plan to further enhance the generalisation of our model to accurately detect more unknown, diverse anomalies. \footnote{Supported by Australian Research Council through grants DP180103232 and FT190100525.}  

\clearpage
%
%
\bibliographystyle{splncs04}
\bibliography{egbib}

\newpage

\beginsupplement
\setcounter{section}{0}

\setcounter{equation}{0}
\setcounter{figure}{0}
\setcounter{table}{0}
\setcounter{page}{1}
\makeatletter
\renewcommand{\theequation}{S\arabic{equation}}
\renewcommand{\thefigure}{S\arabic{figure}}
\newpage

\onecolumn{%
\renewcommand\onecolumn[1][]{#1}%
\begin{center}
    \centering
    \captionsetup{type=figure}
    \includegraphics[width=.95\textwidth]{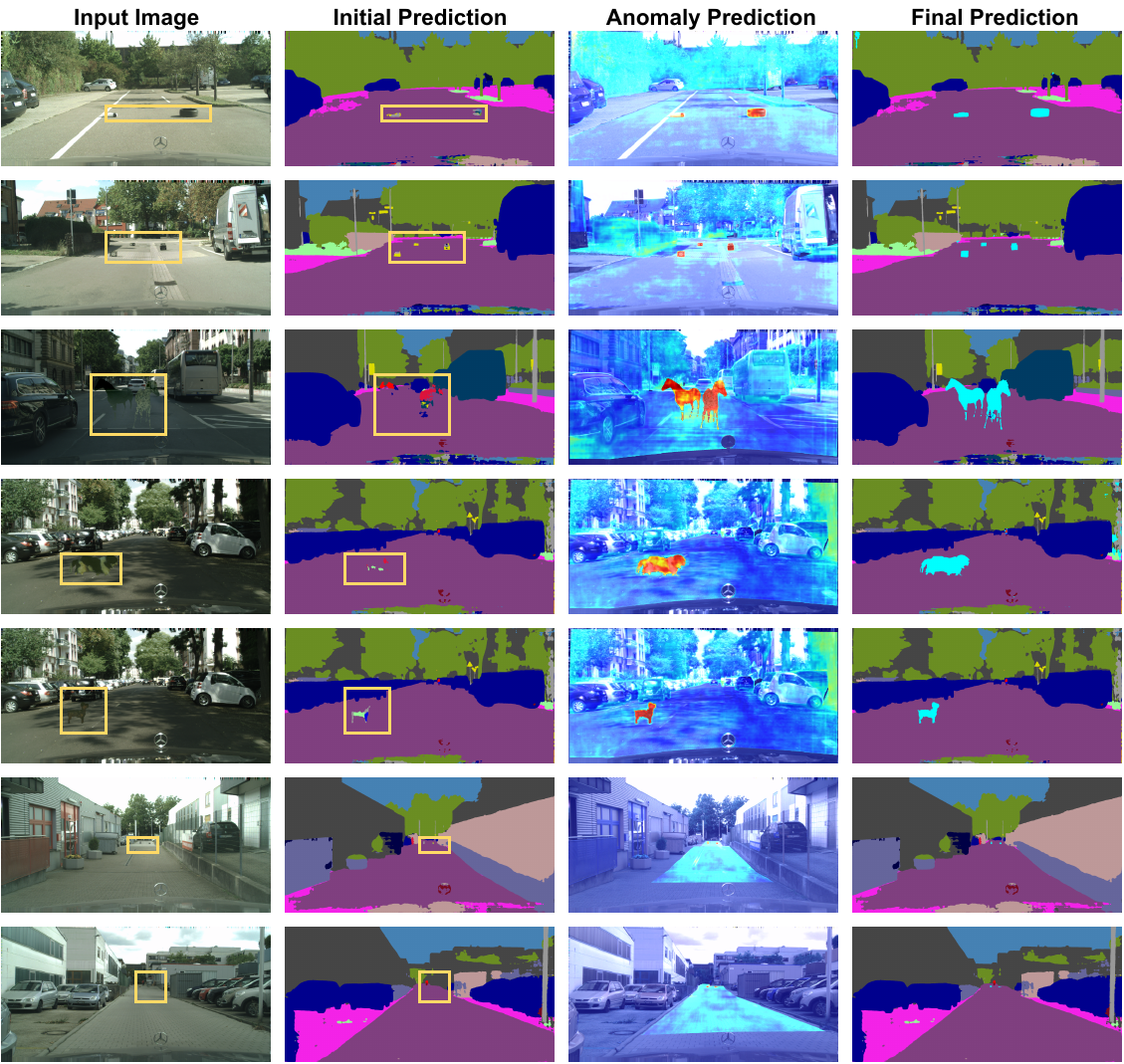}
    \captionof{figure}{From the \textbf{input image} (anomaly highlighted with a yellow box), the \textbf{initial prediction} shows the original segmentation results with anomalies classified as a one of the pre-defined inlier classes.
    \textbf{Anomaly predictions} from \textbf{our} method
    show an anomaly map with high scores (in yellow and red) for anomalous pixels.
    In our \textbf{final prediction}, anomalous pixels are coloured in cyan. 
    }
    \label{fig:supp_img}
\end{center}%
}

\section{Qualitative results}

In Figure~\ref{fig:supp_img}, we show some additional qualitative results. Our approach can effectively detect small and distant objects (rows 6 and 7) and objects with different scales (rows 1 to 5). 

\section{More AUC results}

In Tables~\ref{tab:auc_class_per} and \ref{tab:auc_train_set}, we show the AUC results in addition to the AP and FPR results in Tables 6 and 7 of the main paper.  We achieve consistently SOTA AUC performance regardless of the selection of outlier classes or the number of outlier training samples.

\begin{table}[h]
\centering
\begin{tabular}{@{}c c c@{}}
\toprule 
 Class Per.  & FS LF - AUC  & FS Static - AUC \\ \hline \hline
1\%     & 97.59 {$\pm$0.39}      & 98.37 {$\pm$0.56}\\
5\%     & 98.17 {$\pm$0.45}      & 98.25 {$\pm$0.71}\\
10\%     & 98.47 {$\pm$0.39}   & 99.59 {$\pm$0.25} \\
25\%       & 98.39 {$\pm$0.28}     & 99.52 {$\pm$0.17} \\ 
50\%  & 98.63 {$\pm$0.07}       & 99.54 {$\pm$0.08} \\
75\%   & 98.71 {$\pm$0.05}   &  99.59 {$\pm$0.03} \\ 
 \bottomrule 
\end{tabular}%
\caption{AUC testing results (mean results over six random seeds) of our approach on Fishyscapes benchmark w.r.t. different \textbf{diversity of OE classes}.
}
\label{tab:auc_class_per}
\end{table}

\begin{table}[h]
\centering
\begin{tabular}{@{}c c c@{}}
\toprule
 Train Size  & FS LF - AUC  & FS Static - AUC \\ \hline \hline
5\%     & 98.13 {$\pm$0.12}      & 99.16 {$\pm$0.09}\\
10\%     & 98.35 {$\pm$0.15}   & 99.57 {$\pm$0.07} \\
25\%       & 98.36 {$\pm$0.06}     & 99.51 {$\pm$0.06} \\ 
50\%  & 98.69 {$\pm$0.05}       & 99.37 {$\pm$0.07} \\
 \bottomrule
\end{tabular}%
\caption{{AUC testing results (mean results over six random seeds) of our approach on Fishyscapes benchmark w.r.t. different \textbf{amount of OE training samples}.}
}
\label{tab:auc_train_set}
\end{table}

\section{Hyper-parameters Selection}
For testing, we note a small performance gap with $\lambda \in\{0.1,0.01\}$ 
on LF test set, with AP=78.29 for $\lambda=0.01$ and AP=77.15 for $\lambda=0.1$.  
For the EBM margin, PEBAL reaches AP$\in[76.9,78.3]$
and FPR$\in [0.8,1.3]$ for $m_{in} \in [-12,-22]$ and $m_{out} \in [-2,-8]$  
for different values of $m_{in}$ and $m_{out}$ on LF test set.

\section{Training Details on Cityscapes}
Following~\cite{chan2021entropy,chan2021segmentmeifyoucan}, we use the same DeepLabv3+~\cite{chen2018encoder} with WideResnet38 (90.3 mIoU on Cityscapes Val) trained by Nvidia~\cite{zhu2019improving} as one of the backbones of our segmentation model. 
As mentioned in~\cite{zhu2019improving}, the model is firstly pre-trained on Mapillary Vista dataset~\cite{neuhold2017mapillary}, and then fine-tuned on Cityscapes train set with their proposed label relaxation loss and sdc-aug label propagation. 
Their model uses a different \{cv2: monchengladbach, strasbourg, stuttgart\} validation split than the standard split \{cv0: munster, lindau, frankfurt\}. Please refer to their paper for more details. 
For DeepLabv3+~\cite{chen2018encoder} with Resnet101 backbone (80.3 mIoU on Cityscapes Val) from~\cite{jung2021standardized}, the authors trained their model with the standard cv0 train/validation split using default formulations in~\cite{chen2018encoder}. All those checkpoints are downloaded from their official Github pages.

\section{Results Based on Different DeepLabv3+ Checkpoint}
In this section, we show the results of another DeepLabv3+~\cite{chen2018encoder} with WideResnet38 trained by Nvidia~\cite{zhu2019improving} using the Cityscapes \textbf{\{cv0: munster, lindau, frankfurt\}} standard train/val split. The checkpoint is downloaded from the their official Github page~\cite{zhu2019improving}, with a 81.8\% mIoU on Cityscapes validation set. This model was firstly pre-trained on Mapillary Vista dataset~\cite{neuhold2017mapillary} and then fine-tuned on Cityscapes but without their label relaxation loss and sdc-aug label propagation.  As shown in Tab.~\ref{tab:wres34_fishy_cv0_split}, our model outperforms the previous methods by a large margin on all three benchmarks, regardless of the backbones, the segmentation accuracy and the Cityscapes train/val splits. Notably, our method surpasses the previous SOTA SML by 40\%, 50\% and 20\% of AP on three datasets, respectively. We also achieve best AUC and FPR results on all datasets. 

\begin{table*}[!t]
\centering
\caption{Anomaly segmentation results on \textbf{Fishyscapes validation sets} (LostAndFound and Static), and the \textbf{Road Anomaly testing set}, with \textbf{WideResnet38} backbone under \textbf{cv0} standard train/val split.}
\resizebox{.8\linewidth}{!}{$%
\begin{tabular}{@{}c|ccc|ccc|ccc@{}}
\toprule
\multirow{2}{*}{Methods} & \multicolumn{3}{c|}{FS LostAndFound} & \multicolumn{3}{c|}{FS Static} & \multicolumn{3}{c}{Road Anomaly}  \\ \cline{2-10}
 & AUC $\uparrow$  & AP $\uparrow$  & FPR$_{95}$ $\downarrow$  & AUC $\uparrow$  & AP $\uparrow$  & FPR$_{95}$ $\downarrow$ & AUC $\uparrow$  & AP $\uparrow$  & FPR$_{95}$ $\downarrow$  \\ \hline  \hline
MSP~\cite{hendrycks2019scaling} & 89.26  & 11.84  & 32.55 & 89.26 & 11.84  & 32.55  & 72.37  & 20.23 & 67.98   \\
Max Logit~\cite{hendrycks2019scaling} & 93.14  & 12.78  & 38.15  & 93.27 & 18.89  & 25.49   & 76.39  & 23.46  & 64.55   \\
Entropy~\cite{hendrycks2016baseline} & 89.01  & 8.79  & 47.81  & 90.28  & 15.19  & 31.71  & 73.70 & 22.13  & 67.42   \\
Energy~\cite{liu2020energy} & 93.45  & 14.29 &  37.71 & 93.52  & 19.22  & 25.02  & 76.76   & 23.48   & 64.04 \\ 
SML~\cite{jung2021standardized} & 96.03 	& 21.71 & 20.09	 & 95.79	 &	32.04  & 15.81	 & 74.45	 & 22.16	 & 68.59   \\ \hline
Ours & \textbf{98.52}  & \textbf{64.43}  & \textbf{6.56}  & \textbf{99.33}  & \textbf{86.01}  & \textbf{2.63}  & \textbf{88.85}  & \textbf{44.41}  & \textbf{37.98}  \\ \bottomrule
\end{tabular}%
$} 
\label{tab:wres34_fishy_cv0_split}
\end{table*}

\end{document}